\documentclass[11pt]{article}

\usepackage{natbib}
\usepackage{amsmath,amssymb}
\usepackage[a4paper, marginparwidth=75pt]{geometry} 

\usepackage{caption}
\usepackage{subcaption}
\usepackage{graphicx}
\usepackage{xcolor}
\usepackage{hyperref}
\usepackage{booktabs}

\newcommand{\lang}[1]{\ensuremath{\textit{#1}}}
\newcommand{\vect}[1]{\ensuremath{\textbf{#1}}}
\newcommand{\mat}[1]{\ensuremath{\underline{\vect{#1}}}}

\title{Grounded learning for compositional vector semantics}
\author{Martha Lewis}

\begin{document}

\maketitle

\begin{abstract}
Categorical compositional distributional semantics is an approach to modelling language that combines the success of vector-based models of meaning with the compositional power of formal semantics. However, this approach was developed without an eye to cognitive plausibility. Vector representations of concepts and concept binding are also of interest in cognitive science, and have been proposed as a way of representing concepts within a biologically plausible spiking neural network. This work proposes a way for compositional distributional semantics to be implemented within a spiking neural network architecture, with the potential to address problems in concept binding, and give a small implementation. We also describe a means of training word representations using labelled images.

\end{abstract}

\section{Introduction}
Vector representations of word meaning have proved extremely successful at modelling language in recent years, both as static word embeddings \citep{mikolov2013,pennington2014} and as contextual embeddings which take surrounding words into account \citep{devlin2019}. Vectors have also been used in cognitive science, both at a fairly abstract level representing concepts via collections of features, and at a more mechanistic level representing concepts via patterns of neural activation.

In all cases we have an interest in describing how words or concepts combine. In the case of contextual word embeddings, composition is effected by the artificial neural network architecture, and this works very well, although in an opaque manner and arguably in a way that does not generalise effectively to other tasks \citep{talman2019,bernardy2019} or which leverages specific characteristics of the dataset \citep{mccoy2019a}.

In the case of static word embeddings, compositional distributional semantics describes methods to both build vector representations of words and combine them together so that phrases and sentences can be represented as vectors. \cite{mitchell2010} describe some quite general approaches to composition, and give implementations focussed on pointwise, potentially weighted, combinations, such as vector addition or pointwise multiplication. Grammatically informed neural approaches are given in \citep{socher2013,bowman2015} where artificial neural networks for composing word vectors are built that use the grammatical structure of a sentence. Finally tensor-based approaches were proposed and developed in \cite{coecke2010,baroni2010,paperno2014}. In these approaches words are modelled in different vector spaces depending on their grammatical type, and composition is given by tensor contraction. This will be described in more detail in section \ref{sec:disco}. Compositional distributional semantic approaches are in general used to model text only, although some multi-modal approaches have been used. This leads to the question of whether we can develop a means of learning compositional vector-based representations in a grounded way.

On the cognitive side, we focus here on the idea of vectors as representing patterns of neural activation. One means of considering how vectors combine in this context is given by \emph{vector symbolic architectures} (VSAs) \citep{smolensky1990,plate1994a, gayler2003}. VSAs represent symbols as vectors, and provide a means of binding symbols together, grouping them, and unbinding them as needed. More detail is given in section \ref{sec:vsas}. VSAs have been posited as a way of modelling how symbols can be represented and manipulated in a neural substrate. This has been implemented in e.g. \cite{eliasmith2012} and investigated and discussed in e.g. \cite{hummel2011,martin2020}. In \cite{martin2020} the argument is made that \emph{additive binding}, i.e. combining vectors via addition, is more faithful to how humans combine concepts than \emph{conjunctive binding}, i.e. using something like a tensor product. Since VSAs have been investigated and implemented within more biologically realistic neural networks, the question arises of whether we can use these methods in developing a grounded learning model for tensor-based compositional semantics, all the more so since the model of \cite{coecke2010} was inspired by Smolensky's model originally.

\paragraph{Aims}
\begin{itemize}
	\item To build a model for generating grounded representations within a compositional distributional semantics
	\item To draw out links between Smolensky's theory on one hand and compositional distributional semantics on the other
	\item To develop a way in which compositional distributional models can be implemented within biologically plausible neural network models and thereby investigate a wider range of composition methods than e.g. vector addition or tensor binding.
\end{itemize}

\section{Background}
\subsection{Compositional Distributional Semantics}
\label{sec:disco}
Compositional distributional semantics was developed as a way of generating meanings above the word level via the composition of individual word meanings. The genre of model we concentrate on here can be termed tensor-based compositional distributional models \citep{coecke2010,baroni2010,paperno2014}. Words are modelled in different vector spaces according to their grammatical type, and composition is modelled as tensor contraction. Specifically:
\begin{itemize}
	\item Nouns are modelled as vectors in a noun space $N$, sentences in a sentence space $S$. 
	\item Adjectives are modelled as matrices on $N$, i.e. linear maps $\lang{adj}:N\rightarrow N$ or elements of the space $N \otimes N$. 
	\item Intransitive verbs are modelled as matrices from $N$ to $S$, i.s. linear maps $\lang{iv}:N \rightarrow S$ or elements of the space $N \otimes S$
	\item Transitive verbs are modelled as tensors in $N\otimes S \otimes N$, or bilinear maps $\lang{tv}:N\otimes N \rightarrow S$
\end{itemize}
So, for example, an adjective like red is modelled as a matrix $\mat{red}$ and applied to a noun $\vect{car}$ by matrix multiplication, giving back a vector $\vect{red car}$.

In the original formulation, vectors for words were presumed to be inferred from large text corpora, and so far there have been limited proposals for how to ground these representations using images or other forms of input. Compositional distributional semantics has also been developed with limited consideration of cognitive or neural plausibility. In contrast, vector symbolic architectures, discussed in the following section, have been considered by some cognitive scientists as offering a good basis for modelling language and concept combination.

\subsection{Vector-symbolic architectures}
\label{sec:vsas}
Prior to tensor-based distributional semantics, there has been a large amount of research into vector symbolic architectures (VSAs), specifically, how symbolic structures can be encoded into vector-based representations. These include \cite{smolensky1990,plate1994,gayler2003} amongst others.

\cite{smolensky1990} proposes that structures like sentences are modelled as a sum of role-filler bindings. Suppose we have a set of roles $\{\lang{agent}, \lang{patient}, \lang{verb}\}$ and a set of fillers $\{\lang{Junpa}, \lang{Jen}, \lang{loves}\}$. Symbolically the binding of a role to a filler is represented by $/$, and the sentence \lang{Junpa loves Jen} can be represented as a set of role filler bindings $\{\lang{Junpa}/\lang{agent}, \lang{Jen}/\lang{patient}, \lang{loves}/\lang{verb}\}$. In \cite{smolensky1990} these are mapped over to a vector space model by mapping each role and each filler to a vector, mapping the binding to tensor product, and mapping the collection of the role-filler bindings as their sum:

\begin{align}
	\lang{Junpa loves Jen} &\mapsto \{\lang{Junpa}/\lang{agent}, \lang{Jen}/\lang{patient}, \lang{loves}/\lang{verb}\} \\
	\label{eq:jlj}
	&\mapsto \vect{Junpa}\otimes\vect{agent} + \vect{Jen}\otimes\vect{patient} + \vect{loves}\otimes\vect{verb}
\end{align}

More generally, a sentence $s$ consisting of a set of role-filler bindings $\{r_i/f_i\}_i$ is realized as:
\begin{equation}
	\label{eq:ics}
	\vect{s} = \sum_i \vect{r}_i \otimes \vect{f}_i
\end{equation}

Questions can be asked of a given statement via an \emph{unbinding mechanism}. We may want to extract individual elements of a given sentence. This is done using \emph{unbinding vectors}, defined as vectors dual to the role vectors. Each role vector $\vect{r}_i$ has an unbinding vector $\vect{u}_i$ such that $\langle \vect{r}_i, \vect{u}_i\rangle = 1$. Note that if the role vectors are an orthonormal set, each role vector is its own unbinding vector. To unbind a particular role from a sentence, we take the partial inner product of the unbinding vector with the sentence representation. Suppose that 
\begin{equation}
	\vect{s} = \vect{Junpa}\otimes\vect{agent} + \vect{Jen}\otimes\vect{patient} + \vect{loves}\otimes\vect{verb}
\end{equation}
and $\vect{agent}$, $\vect{patient}$, $\vect{verb}$ form an orthonormal set. If we want to know who the agent is in $\vect{s}$, we take the partial inner product of $\vect{s}$ and $\vect{agent}$ giving:

\begin{align*}
	\vect{s} \cdot \vect{agent} &= (\vect{Junpa}\otimes\vect{agent} + \vect{Jen}\otimes\vect{patient} + \vect{loves}\otimes\vect{verb}) \cdot \vect{agent}\\
	&= \vect{Junpa}\otimes\vect{agent}\cdot \vect{agent} + \vect{Jen}\otimes\vect{patient} \cdot \vect{agent}+ \vect{loves}\otimes\vect{verb} \cdot \vect{agent}\\
	&=\vect{Junpa}
\end{align*}

VSAs have been posited as a potential means for representing symbolic thought in a neural substrate \citep{hummel2011,doumas2012,calmus2020,martin2020}. Holographic reduced representations \citep{plate1994a} have similar properties to Smolensky's theory but without the drawback of needing the increased space for bound representations. \cite{eliasmith2013} have shown how HRRs can be implemented within a spiking neural network model which is designed to be biologically plausible. 

Whilst VSAs such as Smolensky's and Plate's have the benefits outlined above, of being composable and potentially biologically plausible, they have been argued to have some drawbacks in representing how words combine, or how concepts compose. \cite{hummel2011, doumas2012, martin2020} make the following observation. Assuming that similarity is measured by cosine similarity, that is, the cosine of the angle between two vectors, then the similarity of two role-filler bindings is dependent only on the similarity of the pair of roles and the pair of fillers. 

\cite{martin2020} set up an experiment to investigate whether \emph{conjunctive binding} (via tensor product or circular convolution) or \emph{additive binding} (via vector addition) is a better predictor of human similarity judgements. They consider a role to be a predicate, such as \lang{fluffy}, which can be bound to a filler, such as \lang{cat}. Then, the representations for \lang{fluffy dog} and \lang{fluffy cat} are exactly as similar as the representations for \lang{dog} and \lang{cat} are.
\begin{align}
	\vect{fluffy}\otimes\vect{dog}\cdot\vect{fluffy}\otimes\vect{cat} &= \vect{fluffy}\cdot\vect{fluffy} \otimes \vect{dog} \cdot \vect{cat}\\
	&=1 \cdot \vect{dog} \cdot \vect{cat} = \vect{dog} \cdot \vect{cat}
\end{align}

\begin{figure}[htbp]
	\centering
	\begin{minipage}{0.45\textwidth}
		\begin{subfigure}[b]{0.45\linewidth}
			\includegraphics[height=4cm]{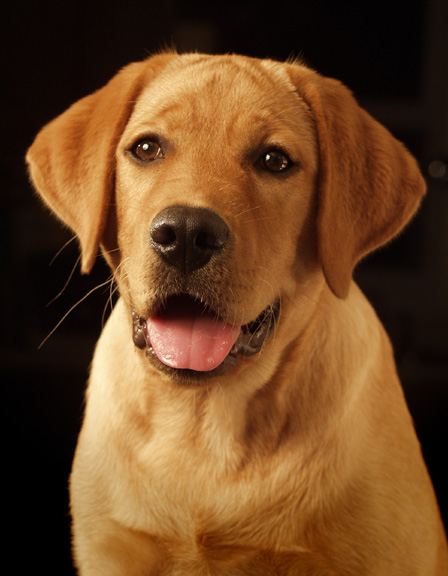}
			\caption{Dog\footnotemark}
			\label{fig:dog}
		\end{subfigure}
		\hfill
		\begin{subfigure}[b]{0.45\linewidth}
			\raggedleft
			\includegraphics[height=4cm]{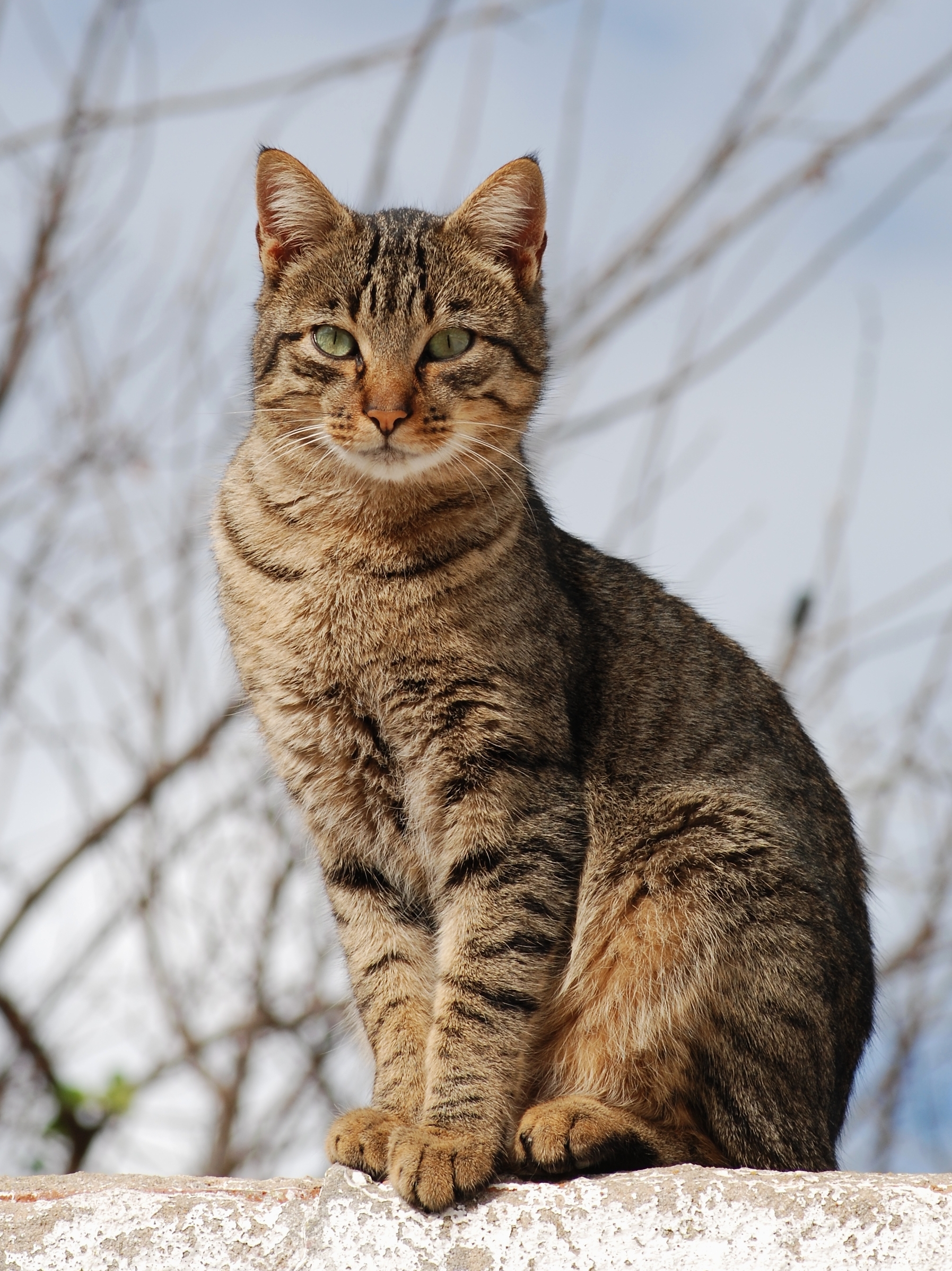}
			\caption{Cat\footnotemark}
			\label{fig:cat}
		\end{subfigure}
	\end{minipage}

	\begin{minipage}{0.45\textwidth}
		\begin{subfigure}[b]{0.45\linewidth}
			\includegraphics[height=4cm]{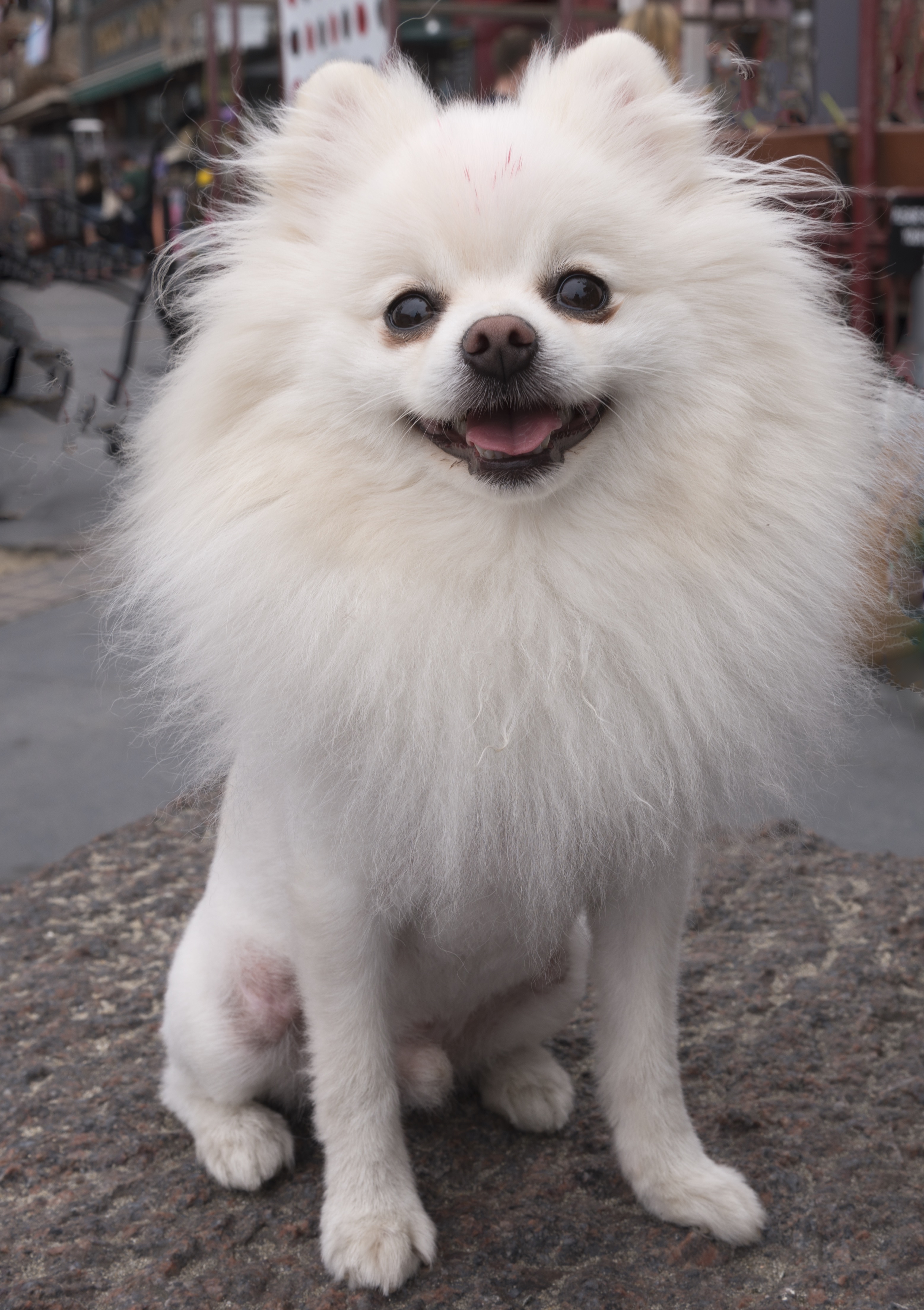}
			\caption{Fluffy Dog\footnotemark}
			\label{fig:fldog}
		\end{subfigure}
		\hfill
		\begin{subfigure}[b]{0.45\linewidth}
			\raggedleft
			\includegraphics[height=4cm]{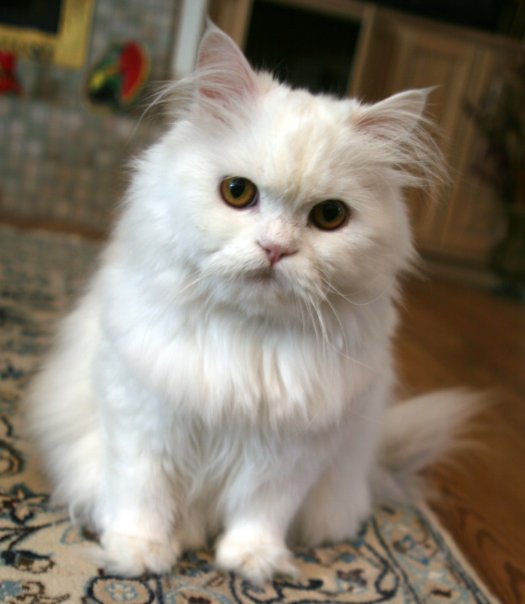}
			\caption{Fluffy Cat\footnotemark}
			\label{fig:flcat}
		\end{subfigure}
	\end{minipage}
	\caption{Fluffy dogs and fluffy cats are more similar than dogs and cats}
	\label{fig:fldc}
	
\end{figure}

This is undesirable, as the predicate fluffy should make cats and dogs more similar to each other - see Figure \ref{fig:fldc} for example.

\addtocounter{footnote}{-4} 
\stepcounter{footnote}\footnotetext{\url{https://commons.wikimedia.org/wiki/File:Wayfield\%27s_Young_Argos,_the_Labrador_Retriever.jpg} 
	
	Attribution: Andrew Skolnick, en:User:Askolnick, CC BY-SA 3.0 <
	\url{http://creativecommons.org/licenses/by-sa/3.0/>, via Wikimedia Commons}}
\stepcounter{footnote}\footnotetext{\url{https://commons.wikimedia.org/wiki/File:Cat_November_2010-1a.jpg} 
	
	Attribution: Alvesgaspar, CC BY-SA 3.0 \url{https://creativecommons.org/licenses/by-sa/3.0}, via Wikimedia Commons}
\stepcounter{footnote}\footnotetext{{\url{https://commons.wikimedia.org/wiki/File:Cute_dog_on_beach,_Los_Angeles,_California_LCCN2013634674.tif}
		
Attribution: Carol M. Highsmith, Public domain, via Wikimedia Commons}}
\stepcounter{footnote}\footnotetext{\url{https://commons.wikimedia.org/wiki/File:White_Persian_Cat.jpg} 
	
	Attribution: Optional at the Persian language Wikipedia, GFDL \url{http://www.gnu.org/copyleft/fdl.html}, via Wikimedia Commons}

Alternatively, roles might be bound to fillers using \emph{additive binding}. In this case, we do obtain the result that fluffy dogs and fluffy cats are more similar than dogs and cats. This property also holds if we model \lang{fluffy} as an adjective, \lang{dog} and \lang{cat} as nouns, and combine the representations as in equation \ref{eq:ics} above. Bearing in mind that we consider $\vect{adj}$ and $\vect{noun}$ to be orthogonal, and assume we have a similarity of around 0.5 for cats and dogs, we have:
\begin{align}
	\lang{fluffy cat}& = \frac{1}{\sqrt{2}}(\vect{fluffy}\otimes\vect{adj} + \vect{cat} \otimes\vect{noun})\\
		\lang{fluffy dog}& = \frac{1}{\sqrt{2}}(\vect{fluffy}\otimes\vect{adj} + \vect{dog} \otimes\vect{noun})
\end{align}
where the factor of $\frac{1}{\sqrt{2}}$ normalises the length of the resulting vectors. Then:
\begin{align}
	sim(\lang{fluffy cat}, \lang{fluffy dog}) &= \frac{1}{2}\langle\vect{fluffy}\otimes\vect{adj} + \vect{cat} \otimes\vect{noun}, \vect{fluffy}\otimes\vect{adj} + \vect{dog} \otimes\vect{noun}\rangle\\
	&= \frac{1}{2}(\langle\vect{fluffy}\otimes\vect{adj}, \vect{fluffy}\otimes\vect{adj}\rangle + \langle\vect{fluffy}\otimes\vect{adj}, \vect{dog} \otimes\vect{noun}\rangle \\
	&\quad + \langle\vect{cat}\otimes\vect{noun}, \vect{fluffy} \otimes\vect{adj}\rangle +  \langle\vect{cat}\otimes\vect{noun}, \vect{dog} \otimes\vect{noun}\rangle)\\
	&= \frac{1}{2}(\langle\vect{fluffy}\otimes\vect{adj}, \vect{fluffy}\otimes\vect{adj}\rangle + \langle\vect{fluffy},\vect{dog}\rangle\langle\vect{adj},\vect{noun}\rangle \\
	&\quad + \langle\vect{fluffy},\vect{cat}\rangle\langle\vect{adj},\vect{noun}\rangle +  \langle\vect{cat}\otimes\vect{noun}, \vect{dog} \otimes\vect{noun}\rangle)\\
	&= \frac{1}{2}(1 + 0 + 0 +  \langle\vect{cat}, \vect{dog}\rangle) = 0.75
\end{align}
so applying \lang{fluffy} has boosted the similarity of cats and dogs.

\paragraph{Why we want more than additive binding} Whilst, as can be seen in \cite{martin2020}, additive binding works for many examples, there are arguably a number of predicates for which we don't want an increase in similarity to occur. Some words are ambiguous, and we do not want to say that \lang{bright light} and \lang{bright student} are more similar than \lang{light} and \lang{student} are. One argument might be that we should disambiguate words and that these different senses will be represented by different vectors. However, some ambiguities can be subtle: compare \lang{Nishi opened the book} with \lang{Nishi opened the jar}.

\cite{boleda2020} argues that tensor-based compositional distributional models of meaning can reflect polysemy, and this assertion is borne out by the experimental results of \cite{grefenstette2011}, where a matrix-based method is compared with (amongst others) an additive model. A hand-crafted example of this is given in \cite{grefenstette2010} where they show how to design a representation for \emph{catch} so that the similarity of the phrases \emph{catch ball} and \emph{catch disease} is not boosted by the word \emph{catch}. \cite{coecke2015} look at the pet fish problem, and show how a representation for \emph{pet} can be developed that sends \emph{fish} to \emph{goldfish} but leaves \emph{dog} and \emph{cat} mostly unchanged.

The compositional distributional model is focussed on corpus-based semantics, and was not developed with any eye to neural plausibility or to learning language in a grounded fashion. We would like to be able to combine the greater flexibility of tensor-based compositional models with the neural plausibility of VSA-based architectures, and with the possibility of learning meanings that are grounded in an input outside of text. We provide a mapping from the compositional distributional model into a Smolensky-based architecture. We show how meanings of words can be learnt in a way that is grounded in inputs outside of text.

\paragraph{Relations to quantum models of concepts} The compositional distributional semantic model of meaning proposed in \cite{coecke2010} has its roots in quantum theory. There is a wide range of interest in quantum modelling of concepts, examining how concepts behave in composition, and their application in artificial intelligence. \cite{widdows2021} provides a thorough review of applications of quantum theory to artificial intelligence, part of which is to do with the representation of concepts or words by vectors, together with an analysis of the use of VSA-like architectures that we cover here.  In the area of concept composition, it is proposed that concept composition can be well modelled as interference between two quantum states \citep{aerts2009,aerts2012}. Aspects such as emergent meaning and vagueness are addressed in \cite{bruza2012,blutner2013}, and reviews of the use of quantum theory in cognitive science are given in \cite{pothos2013} and \cite{lewis2021}. Further consideration of whether these phenomena can be well modelled within neural networks is an area of future work.

\subsection{Semantic Pointers for Concept Representation}
Whilst deep neural architectures have had huge success in recent years, they are biologically implausible in the structure of the individual units, the overall architecture of networks, and in the learning algorithm implemented. There has therefore been interest in implementing more biologically plausible networks. One of these is Nengo \citep{eliasmith2012}. Within this architecture, symbolic structure is implemented using the Semantic Pointer Architecture (SPA) \citep{eliasmith2012,blouw2016}. Semantic pointers can be thought of as vectors that are instantiated by patterns of neurons firing in a spiking neural network. Semantic pointers can be bound together using circular convolution, and unbound using circular correlation \citep{plate1994}. A Smolensky-style form of concept composition can readily be implemented - with the caveat that these representations and binding and unbinding operations are noisy.

In what follows, we propose a way to view tensor-based compositional distributional semantics within a Smolensky-style framework, and thereby propose a way for tensor-based compositional distributional semantics to be implemented within a spiking neural network architecture.

\section{Compositional distributional semantics in the Nengo framework}
The Nengo framework uses the Semantic Pointer Architecture to represent concepts. Semantic pointers can be thought of as noisy vectors encoded by the dynamics of  a spiking neural network. Semantic pointers are bound together using circular convolution, and can be unbound using circular correlation \citep{plate1994}. A usual proposal for the representation of concepts in this kind of framework is to consider features of a concept as roles, the value of those features as fillers, and form the representation of a concept as a sum of role-filler binding. However, this then leads to the question of how function words like adjectives and verbs should be represented, and how they might be combined with noun concepts.

\subsection{A first proposal}
Recall that a key aspect of vector symbolic architectures is the existence of a binding operator and an unbinding operator. In Smolenksy's ICS these are respectively tensor product and inner product, and in \cite{plate1994} these are circular convolution and circular correlation.

In compositional distributional semantics, we also make use of the inner product (or more generally tensor contraction) as a composition operator, and function words such as verbs and adjectives are tensors or matrices, i.e. weighted sums of tensor products of basis vectors. 
\textbf{We therefore have an immediate way of mapping to the semantic pointer architecture needed for implementation in Nengo}, by viewing inner product as an unbinding operator and tensor product as a binding operator.

In a little more detail: given a matrix $\underline{\vect{red}}$ and a vector $\vect{car}$, we form the composition $\vect{red car}$ via matrix multiplication. Writing this out explicitly, if we have a noun space $N$ with basis $\{\vect{e}_i\}_i$, then 
\begin{align}
    \vect{car} &= \sum_i c_i \vect{e}_i,\qquad
    \underline{\vect{red}}= \sum_i r_{ij} \vect{e}_i \otimes \vect{e}_j\\
    \vect{red car} &= \sum_{ij}r_{ij} \vect{e}_i \langle \vect{e}_j, \vect{car}\rangle = \sum_{ijk}r_{ij} c_k \vect{e}_i \langle \vect{e}_j, \vect{e}_k\rangle = \sum_{ij}r_{ij} c_j \vect{e}_i
\end{align}
i.e., we can think of this operation as unbinding the vector $\vect{car}$ from the adjective $\underline{\vect{red}}$.

Now, to move to the semantic pointer setting, we map tensor product to circular convolution, and inner product to circular correlation. We view each basis vector as a semantic pointer, and encode a noun as a weighted sum of semantic pointers, and an adjective as a weighted sum of convolved pairs of semantic pointers:

\begin{align}
\label{eq:spa1}
    \vect{car} &= \sum_i c_i \vect{p}_i, \qquad
    \underline{\vect{red}}= \sum_i r_{ij} \vect{p}_i \circledast \vect{p}_j\\
    \label{eq:spa2}
    \vect{red car} &= \sum_{ij}r_{ij} \vect{p}_i ( \vect{p}_j \oslash\vect{car})  = \sum_{ijk}r_{ij} c_k \vect{p}_i ( \vect{p}_j  \oslash \vect{p}_k)  = \sum_{ij}r_{ij} c_j \vect{p}_i + \text{noise}
\end{align}
The last step in the above relies on the semantic pointers $\vect{p}_i$ being approximately orthogonal, which they are by design.

A toy implementation of this is available at \url{https://github.com/marthaflinderslewis/nengo-disco}. We use the `pet fish' problem as an example. In the pet fish problem, we want the adjective `pet' to modify animals in certain ways. A `pet fish' should modify `fish' to make it similar to a goldfish, however, the representation of `cat', and `dog' should stay pretty similar: cats and dogs are already pretty archetypal pets. To implement a model of the `pet fish' problem, we take inspiration from \cite{coecke2015}. We choose some features to describe our animals: \lang{cared-for}, \lang{vicious}, \lang{fluffy}, \lang{scaly}, \lang{lives-house}, \lang{lives-sea}, \lang{lives-zoo}. These are rendered as semantic pointers and we use the following notation: \lang{cared-for}: $\vect{c}$, \lang{vicious}: $\vect{v}$, \lang{fluffy}: $\vect{f}$, \lang{scaly}: $\vect{s}$, \lang{lives-house}: $\vect{h}$, \lang{lives-sea}: $\vect{e}$, \lang{lives-zoo}: $\vect{z}$. Each animal is rendered as a weighted sum of semantic pointers with weights as in table  \ref{tab:sps}. We interpret these weights as the importance of each feature to the noun. Note that vectors are normalised.

\begin{table}[h]
    \centering
    \begin{tabular}{c|cccccc}
                 & Fish & Goldfish  & Cat   & Dog   & Shark & Lion \\ \midrule
         \vect{c}& 0.13 & 0.44      & 0.57  & 0.67  & 0.00  & 0.19 \\
         \vect{v}& 0.51 & 0.00      & 0.13  & 0.37  & 0.57  & 0.62 \\
         \vect{f}& 0.00 & 0.00      & 0.57  & 0.37  & 0.00  & 0.44 \\
         \vect{s}& 0.63 & 0.62      & 0.00  & 0.00  & 0.57  & 0.00\\
         \vect{e}& 0.51 & 0.00      & 0.00  & 0.00  & 0.57  &0.00 \\
         \vect{h}& 0.19 & 0.62      & 0.57  & 0.52  & 0.00  & 0.00 \\
         \vect{z}& 0.19 & 0.19      & 0.00  & 0.00  & 0.11  & 0.62
    \end{tabular}
    \caption{Weights for semantic pointer representations of nouns}
    \label{tab:sps}
\end{table}
We then design an adjective as the following sum of convolved semantic pointers:
\begin{align}
    \underline{\lang{pet}} = \vect{c} \circledast (\vect{c} +\vect{v} + \vect{f} + \vect{s}+ \vect{e} + \vect{h} + \vect{z}) + \vect{v}\circledast \vect{v} + \vect{f}\circledast \vect{f}+\vect{s}\circledast \vect{s} + \vect{h}\circledast(\vect{h}+\vect{e}+\vect{z})
\end{align}
which in matrix format looks as in table \ref{tab:adj}
\begin{table}[h!]
    \centering
    \begin{tabular}{c|ccccccc}
                 & \vect{c} & \vect{v}  &\vect{f}   & \vect{s}   & \vect{e} & \vect{h} & \vect{z}\\ \midrule
         \vect{c}& 1 & 1      & 1  & 1  & 1  & 1 & 1 \\
         \vect{v}& 0 & 1 & 0 & 0 & 0& 0& 0\\
         \vect{f}& 0 & 0 & 1 & 0 & 0& 0& 0 \\
         \vect{s}& 0 & 0 & 0 & 1 & 0& 0& 0\\
         \vect{e}& 0 & 0 & 0 & 0 & 0& 0& 0 \\
         \vect{h}& 0 & 0 & 0 & 0 & 1& 1& 1 \\
         \vect{z}& 0 & 0 & 0 & 0 & 0& 0& 0
    \end{tabular}
    \caption{Weights for sum of convolved semantic pointers}
    \label{tab:adj}
\end{table}
We can interpret these weights as follows. The first row of the matrix is essentially saying that no matter what the features of the animal, after application of the \lang{pet} adjective, the animal should be cared for. The next three rows are just identity. The rows corresponding to \lang{lives-sea} and \lang{lives-zoo} are zero: after application of the \lang{pet} adjective, the pet animal should not have any weight on these features. Lastly, we see that the row corresponding to \lang{lives-home} moves weight from other features to this feature.

In Nengo, the nouns and adjective are implemented as weighted sums of semantic pointers or convolved semantic pointers, and the nouns and adjective are composed using the unbinding mechanism: the nouns are unbound from the adjective. Each adjective-noun combination is then queried against the nouns to retrieve the noun that is most similar. We wish that `pet fish' is most similar to `goldfish', `pet cat' is most similar to `cat', and so on. A video of the system can be seen at \url{https://github.com/marthaflinderslewis/nengo-disco}.

The above goes to show that compositional distributional semantics can be implemented within the semantic pointer architecture, but does not give any indication about how features or weights could be learnt. In the following section, we give an alternative formulation which has the potential to provide a learning mechanism from labelled stimuli. 

\subsection{Compositional distributional semantics as a role-filler model of meaning}
We now provide a slightly different perspective on compositional distributional semantics within a semantic pointer architecture.
As we described in section \ref{sec:vsas}, equation \eqref{eq:jlj}, a semantic representation in ICS consists of a sum of role-filler pairs:
\begin{equation}
	\vect{s} = \sum_i \vect{r}_i \otimes \vect{f}_i
\end{equation}
In order to map this representation to the compositional vector representation we consider the following. In the case of a noun, we say that the roles $\vect{r}_i$ are a set of basis vectors spanning the noun space, and then the fillers are simply scalars attached to each role.
\begin{equation}
	\vect{n} = \sum_i n_i \vect{r}_i
\end{equation}

We view an adjective as a set of fillers bound to roles where the roles are possible nouns and the fillers are vectors corresponding to the adjective-noun combination:
\begin{equation}
	\label{eq:adjs}
	\vect{adj} = \sum_i \vect{an}_i \otimes \vect{n}_i
\end{equation}

We view intransitive verbs as a set of fillers bound to roles where the roles are possible nouns and the fillers are the resulting sentences:
\begin{equation}
	\vect{in-verb} = \sum_i \vect{n}_i \otimes \vect{sent}_i
\end{equation}

Transitive verbs are a set of fillers bound to roles where the roles are pairs of possible nouns and the fillers are the resulting sentences:
\begin{equation}
	\vect{tr-verb} = \sum_{ij} \vect{n}_i \otimes \vect{sent}_{ij} \otimes \vect{n}_j
\end{equation}

The composition of an adjective and a noun is then found by unbinding the noun role from the adjective, and similarly for verb-noun composition. In the adjective-noun example, unbinding is just matrix-vector multiplication.
For a very toy example, suppose we have some kind of vector representations of: $\lang{red car} = \vect{rc}$, $\lang{red apple}= \vect{ra}$, $\lang{red wine}= \vect{rw}$, $\lang{car}= \vect{c}$, $\lang{apple}= \vect{a}$, $\lang{wine}= \vect{w}$

Then, 
\begin{align}
	\underline{\vect{red}} = \vect{rc}\otimes\vect{c} + \vect{ra}\otimes\vect{a} + \vect{rw}\otimes\vect{w}
\end{align}
Computing $\lang{red car}$ as $\underline{\vect{red}}\cdot \vect{c}$ we obtain:

\begin{align}
	\underline{\vect{red}}\cdot \vect{c} &= \vect{rc}\langle \vect{c}, \vect{c}\rangle + \vect{ra} \langle\vect{a}, \vect{c}\rangle + \vect{rw} \langle \vect{w}, \vect{c}\rangle\\
	&= \vect{rc} +\text{noise}
\end{align}
assuming that cars are not very similar to apples or wine. 

Recall that in equations \eqref{eq:spa1} and \eqref{eq:spa2} we argued that in order to implement a distributional semantic model within the spiking neural architecture, we could map from tensor product as binding operator and inner product as unbinding operator, to circular convolution as binding operator and circular correlation as unbinding operator. We use the same methodology to obtain a representation of nouns, adjectives, and verbs within the semantic pointer architecture. For example, in the semantic pointer architecture,
\begin{align}
	\underline{\vect{red}} = \vect{rc}\circledast\vect{c} + \vect{ra}\circledast\vect{a} + \vect{rw}\circledast\vect{w}
\end{align}
and
\begin{align}
	\underline{\vect{red}}\oslash \vect{c} &= \vect{rc}( \vect{c}\oslash\vect{c}) + \vect{ra}(\vect{a} \oslash \vect{c}) + \vect{rw} ( \vect{w}\oslash \vect{c})\\
	&= \vect{rc} +\text{more noise}
\end{align}
again, assuming that cars are not very similar to apples or wine. 

This perspective on compositional distributional semantics as a role-filler binding architecture also gives us a potential way of learning representations for words.

\subsection{Learning strategy}
We consider adjective-noun composition. Suppose we have a set of images labelled with adjective-noun combinations, and assume that within our semantic pointer architecture we have some kind of vision system that can produce semantic pointers for the images themselves. 

\paragraph{Supervised learning situation} We have a set of labelled inputs. Let's assume they are all of the form adj-noun. Suppose the system has a convolved semantic pointer representation of each adjective it has learnt so far, call them $\{\mat{A}_i\}_i$, and a semantic pointer representation of each noun it has learnt so far, call them $\{\vect{n}_i\}_i$. Suppose an input is labelled $A_jn_k$ and the system has both these words in its vocabulary. We assume the system has a vector representation $\vect{an}_i$ of each image.

If we assume that the adjective has the form $\vect{adj} = \sum_i \vect{an}_i \circledast \vect{n}_i$ then we get a simple update rule for the adjective, by mixing the current adjective with the convolution of $\vect{an}$ and $\vect{n}$: 
\begin{equation}
	\mat{A}_j \mapsto (1-h)\mat{A}_j + h \vect{an}_i \circledast \vect{n}_k 
\end{equation}
where $h$ is some small value in $[0, 1]$. In order to update the noun, we propose  unbinding the $\vect{an}$ filler from $\mat{A}$ to get the $\vect{n}$ role, giving us 
\begin{equation}
	\vect{n}_k \mapsto (1-h)\vect{n}_k + h(\vect{an}_i \oslash \mat{A}_j)
	\label{eq:update_A}
\end{equation}

In the cases where the system does not yet have a representation of the adjective, we can initialise it as $\vect{an} \otimes \vect{an}$ and where there is no representation of the noun, it can be initialised as $\vect{an}$.

We can make a similar proposal for intransitive verbs. We assume that the verb has the form $\vect{verb} = \sum_i \vect{n}_i \circledast \vect{nv}_i$, where $\vect{nv}$ is a vectors representing an intransitive sentence like `Junpa walks'. Given an input labelled $n_jV_k$ corresponding to a vector representation $\vect{nv}_i$, the verb $V_k = \sum_i \vect{n}_i \circledast \vect{nv}_i$ is then updated by:
\begin{equation}
	\mat{V}_k \mapsto (1-h)\mat{V}_k + h \vect{n}_j \circledast \vect{nv}_i 
\end{equation}
and the noun is updated by:
\begin{equation}
	\vect{n}_j \mapsto (1-h)\vect{n}_j + h(\mat{V}_k \oslash \vect{nv}_i)
	\label{eq:update_n}
\end{equation}

We have started to investigate the approach outlined above within a standard neural network model to learn compositional word representations from labelled images \citep{lewis2023}, and future work will go on to extend this to the Nengo architecture.

\section{Conclusions and future work}
Compositional distributional semantics has a set of powerful machinery that can be used for composition. However, it does not have any particular cognitive grounding. In this paper, we have given a proposal for implementation of compositional distributional semantics within the cognitive architecture Nengo. This architecture uses a biologically (more) realistic substrate to represent concepts as semantic pointers, and is integrated with decision making, vision, and other modules. This therefore has the potential to provide compositional distributional semantics with an environment in which meanings can be grounded. We have given a toy implementation to show that our proposal is possible, and presented an alternative formulation of the approach together with a strategy for learning word representations. 

Future work in this area is of course to take forward possible implementations of these ideas. We already have a strategy to begin implementation within the Nengo framework. Once implementation within this kind of architecture has been carried out, there is potential to examine what kind of representation best model human behaviour - whether compositional distributional semantics is a useful representation. We gave arguments in section \ref{sec:vsas} to argue that there is a need for more flexible composition than additive binding, so there is potential here.

We would also like to implement these ideas within a dialogue setting. Whilst a supervised learning setting has been described, the kinds of representation proposed are very amenable to being learnt in a self-supervised fashion between two or more agents.

\bibliographystyle{plainnat}
\bibliography{neural_tensors}

\end{document}